# A review of faithfulness metrics for hallucination assessment in Large Language Models

Ben Malin, *Member, IEEE*, Tatiana Kalganova, *Member, IEEE* and Nikolaos Boulgouris, *Member, IEEE*

*Abstract*— **This review examines the means with which faithfulness has been evaluated across open-ended summarization, question-answering and machine translation tasks. We find that the use of LLMs as a faithfulness evaluator is commonly the metric that is most highly correlated with human judgement. The means with which other studies have mitigated hallucinations is discussed, with both retrieval augmented generation (RAG) and prompting framework approaches having been linked with superior faithfulness, whilst other recommendations for mitigation are provided. Research into faithfulness is integral to the continued widespread use of LLMs, as unfaithful responses can pose major risks to many areas whereby LLMs would otherwise be suitable. Furthermore, evaluating open-ended generation provides a more comprehensive measure of LLM performance than commonly used multiple-choice benchmarking, which can help in advancing the trust that can be placed within LLMs.**

## I. INTRODUCTION

Generative AI has had significant recent advancements and can be used to produce text, images or many different types of output [1]. Furthermore, it can serve as the engine in various intelligent applications, the most notable of which is Large Language Models (LLMs). LLMs have become integral to numerous domains, leveraging their ability to understand and generate human language at a remarkable scale. These models, such as ChatGPT, BERT, and T5, have been applied across various fields, significantly improving natural language understanding and generation [1]. Despite their impressive applications, LLMs are not always reliable, and they tend to hallucinate, i.e., occasionally produce responses that are clearly wrong or non-factual. These hallucinations necessitate the development of metrics and benchmarks that can assess the quality of LLM response to help detect and minimise hallucination. Through identification of these hallucinations allows for their mitigation in a variety of manners, highlighting the importance of reliable identification. LLMs can be applied in almost any field, including safety critical applications, whereby the production of hallucinatory responses can result in catastrophic outcomes. The discourse relating to hallucinations can be further broadened into the evaluation of faithfulness, which more specifically relates to the identification of generated outputs which are unfaithful to the desired output. In the present review article, we discuss existing methodologies for assessing LLM operations in terms of their ability to perform tasks such as translation, summarization and question answering. We present analysis regarding datasets and metrics that can be used for evaluating faithfulness across domains, as well as addressing gaps within the literature where novel metrics can be applied. Finally, the means to mitigate hallucinations, and improve faithfulness in generated outputs are assessed.

This review follows the Introduction with Section 2 which describes the importance of faithfulness for LLMs as well as the diverse applications they can be used for. Section 3 details commonplace metrics across all of the domains within the scope of this review. Section 4 segments the domains and provides analysis into the datasets used for each domain, and their unique requirements for the assessment of faithfulness. Section 5 assesses the findings that various studies have gathered regarding the capabilities of different metrics in each domain. Finally, section 6 provides discussion into the compilation of these findings for the mitigation of hallucinations as well as current potential areas for research within assessing faithfulness.

## II. FAITHFULNESS IN LLMS

### A) Hallucinations

In the context of Large Language Models (LLMs), faithfulness refers to the degree to which the generated output remains accurate, reliable, and aligned with the ground truth or source material. It is common for LLM-generated outputs to contain "hallucinations", where hallucinations are instances of the LLM introducing incorrect statements, either through misattribution of a fact, or more commonly through inventing one completely. It has been theorized that they are resultant from mismatching data within training, but Parikh states that it is inevitable even with this divergence resolved, as well as positing that it is exacerbated due to text duplication [2]. Whilst Ji states that there is innate divergence due to the sources of textual data that LLMs are trained on [3]. If the data that LLMs are trained on will always result in the possibility of hallucinatory outputs then the ability to identify, and thus

This work has been funded by the European Union. Views and opinions expressed are however those of the authors only and do not necessarily reflect those of the European Union or European Commission-EU. Neither the European Union nor the granting authority can be held responsible for them.

Ben Malin is with Brunel University London, London, U.K. (e-mail: ben.malin@brunel.ac.uk)

Tatiana Kalganova is with Brunel University London, London, U.K. (e-mail: Tatiana.kalganova@brunel.ac.uk).

Nikolaos Boulgouris is with Brunel University London, London, U.K. (e-mail: nikoloas.boulgouris@brunel.ac.uk)

mitigate these erroneous outputs is fundamental to the implementation of generative AI in domains where fictitious responses are unacceptable.

*B) Applications*

Due to the broadness of tasks with which LLMs are often used for, there are several domains which have utilized unique strategies for assessing faithfulness, with these methodologies having been specifically chosen and optimised for the domain at hand. Faithfulness is a crucial evaluation criterion for all domains within the scope of this review, as users rely on the model to provide information that is factually correct and true to the original source otherwise problems will arise with varying levels of severity depending on the criticality of the task at hand.

It can be challenging to define faithfulness universally due to the intricacies required across domains. For example, summarization requires the output to be aligned with the source material, which can be validated without the requirement of a ground truth. Whereas an open question-answering (QA) task requires a ground truth to be provided for the faithfulness to be measured.

The domains that are covered within this review are summarization, question-answering and machine translation. In summarization, LLMs are used to condense large texts into concise summaries, offering practical applications in news aggregation, academic research, and legal documentation. Question-answering LLMs are used to retrieve and generate precise answers from vast datasets, powering search engines, customer service platforms, and knowledge systems Finally, machine translation LLMs are used to produce semantically equivalent text in alternative languages, empowering the sharing of knowledge. Each of these applications, to varying levels of incurred burden, highlights the necessity of evaluating the faithfulness of the generated content, ensuring that outputs remain accurate and aligned with the source information.

### III. Metrics

Faithfulness is often more difficult to measure than other criteria like fluency or grammatical correctness because it involves aligning both the factual content and the semantic meaning, where semantic meaning can be incredibly challenging to automatically evaluate. Ensuring faithfulness is critical for preventing misinformation, ensuring reliability in user-facing applications, and maintaining trust in AI-driven systems, especially in knowledge-intensive or sensitive domains.

This review focuses on evaluating automated metrics designed to assess the faithfulness of outputs generated by LLMs across a range of domains, including summarization, QA and machine translation. Each domain's ubiquitous and automated metrics for assessing faithfulness, such as ROUGE [4], BLEU [5] and BERTScore [6], are discussed in addition to lesser-used metrics that may have only been used in specific domains and assessing their potential use case in alternative domains.

Additionally, this review contrasts these automated metrics with human evaluation, which remains the gold standard for assessing faithfulness. Human evaluators can assess not only factual correctness but also nuanced meaning, context, and the intent behind model outputs. However, human evaluation is resource-intensive, time-consuming, and prone to subjective variation, making automated metrics an appealing alternative despite their limitations.

Aiming to highlight areas where automated metrics still fall short when compared to human judgment, outlining potential avenues for improving the robustness and accuracy of faithfulness evaluation in LLMs. These metrics are assessed across the following domains: summarization, whereby a source document, text passage or dialogue is summarized concisely; open question-answering, specific answers are required by open-ended questions and can be of varying lengths, typically either entity-based or full sentences[7]; machine translation, the conversion of text from a source language into a target language.

*A) Overview of evaluation metrics*

Evaluation metrics for assessing the performance of language models can be broadly categorized into automatic and human evaluation. Automatic evaluation metrics, such as ROUGE, BLEU, BERTScore, and fact-based approaches, are computational methods that quickly assess aspects like fluency, faithfulness, or semantic similarity without human involvement. They are scalable and efficient but may miss nuanced errors or factual inconsistencies. Human evaluation, on the other hand, involves human annotators judging the quality of generated outputs based on criteria like relevance, factual correctness, and coherence. While human evaluation is more reliable in capturing subjective and complex dimensions, it is resource-intensive, time-consuming, and prone to subjective variation. Human evaluation is very commonly used to validate the success of automatic metrics, whereby high Pearson correlation coefficients show that an automatic metric makes similar judgements to a human evaluator.

*B) Faithfulness Metrics for open-ended text generation*

Open-ended text generation contrasts multiple-choice metrics, through the requirement of unstructured text outputs. This makes the evaluation of faithfulness more challenging due to the ambiguity of what makes an output faithful. These metrics are more aligned with real-world scenarios, as it is commonplace for LLMs to be tasked with the production of open-ended text.

*N-grams*

N-gram metrics are commonly used in natural language processing to evaluate the similarity between a generated text and a reference text by comparing overlapping sequences of words (n-grams). An n-gram refers to a contiguous sequence of "n" words or tokens. For example, a 1-gram (unigram) is a single word, a 2-gram (bigram) is a sequence of two words, and so on. N-gram metrics work by breaking down both the generated and reference texts into n-grams, then calculating the level of overlap. The higher the overlap, the more similar the texts are deemed.

The ROUGE (Recall-Oriented Understudy for Gisting Evaluation) family of metrics is widely used for evaluating a wide variety of text generation tasks, likely due to its low computational requirements. However, these metrics do not consider synonyms or semantic understanding and can thus be too strict to reliably measure textual similarity. In addition to this, these approaches weigh all text equally (such as the use of connective words), even though a small number of n-grams convey the majority of semantic information [8].

These issues regarding n-gram metrics have led to numerous authors stating that these approaches are insufficient at measuring the factual correctness of summaries and attain a poor correlation with human factuality judgements [9-11]. Despite these well-known issues regarding n-gram metrics, it remains popular among most domains of LLM faithfulness evaluation.

*Exact match (EM)*

Exact match (EM) as a metric is comparable to n-gram metrics, yet stricter, as it identifies whether the reference text exactly matches the source/ground truth text. Thus, the output of this computation is binary, and offers no flexibility, with several authors outlining such limitations [7,12]. These factors make this metric very useful for some domains, such as multiple-choice QA, or entity-based QA where only short, concise, unambiguous answers are required.

This rigidity of EM can be lessened to a degree through the implementation of a pool of valid answers, allowing for synonyms or aliases to still be deemed correct. However, this will incur additional burden on data collection. For more open-ended, ambiguous, or lengthier text generation tasks this metric is unlikely to be useful. For example, in the abstract summarization domain there will never be a single correct answer, and thus EM is unlikely to be able to evaluate an output as faithful, even if other metrics would deem it such [7]. However, when the desired output is simple, such as Yes/No responses, EM can often be the optimal approach due to the lack of flexibility no longer being an issue [13].

*Lexical match (LM)*

This metric is comparable to EM, but with less stringent requirements. Typically, only requiring the gold answer to appear within the output to be deemed as faithful. This approach mitigates some of the false negatives that can arise due to LLMs providing superfluous levels of detail when responding to a prompt.

Lexical matching is only typically used within the QA domain, due to the inherent binary nature of faithfulness within that area. As opposed to summarization, whereby there is greater ambiguity as to whether a summary conveys and entails the relevant information from a source. Several studies have utilized lexical matching as opposed to EM due to these benefits, and state better suitability for QA tasks [7,14].

*Question Generation*

Within the domain of abstract summarization, question generation has been widely used as a measure of faithfulness [8,15,16,17]. This methodology involves the generation of questions from both the model-generated summary and the corresponding source document, then the answers to these questions are compared to assess factual consistency. If the answers to the questions generated from the summary align with those from the reference, the summary is considered faithful to the original text. This metric can output a score from 0-1, that shows the percentage of questions that had aligned answers (whereby a score of 1 demonstrates that every answer provided by the summary is in alignment with the answer provided by the source document) [8, 16].

Various authors have developed and implemented their own question-generation-answering-based metrics, such as FEQA, QAGS, $Q^2$ and QuestEval [8,15,16,17]. However, they all suffer from being dependent upon both the question-generation model, as well as the question-answering model. As, if the generated questions are of low quality, such as through wording, ambiguity, or sparseness, then faithfulness is unlikely to be able to be evaluated reliably. Likewise, the performance is also dependent upon the question-answering model, as questions can be incorrectly answered despite adequate context being provided, which can lead to unreliable faithfulness evaluation.

These QA-based metrics are also focused on extractive information, whereby questions can be answered directly through available text, making them less well-suited for tasks requiring more abstractive reasoning. Finally, these techniques also incur major computational overheads due to the additional models that are necessitated for both question-generation and question-answering.

*Embedding-based*

Embedding-based metrics are evaluation methods that assess the similarity between generated text and reference text through comparing their semantic representations in a vector space. These metrics typically utilize pre-trained language models to generate embeddings for textual data, capturing deeper semantic relationships between the texts than can be captured with n-gram methodologies.

Given both reference text and outputted text, an embedding-based evaluation would convert the tokens within both documents into vectors, using a language model. The similarity between these embeddings can then be assessed using various computational measures. BERTScore is a commonplace metric, whereby the BERT language model generates the token embeddings for both texts, then the cosine similarity is calculated between both sets of embeddings to provide a similarity score. However, other similarity calculations exist, such as the MoverScore or Word Mover's Distance (WMD) [18,19]. These metrics measure similarity through calculating the minimum distance embedded words from one document would need to move to occupy the vector space inhabited by the embedded words from the opposing document. Furthermore, they show WMD to outperform BERTScore in the majority of tasks [18]. However, despite many authors stating superior performance from alternative embedding-based approaches, BERTScore is the most ubiquitous [18,20]. Additionally, more modern token embedding models have been published, that can offer superior performance to BERT.

Embedding-based methodologies do have limitations, most notably regarding semantically similar tokens that do not convey the same fact. This can lead to issues when identifying faithfulness as the embeddings for sentences may be highly similar to completely different sentences just because of similar subject matter [16].

*Fact extraction*

Fact extraction typically uses computational means to identify facts from text and allows for faithfulness to be assessed based on core factual elements rather than surface-level similarities, making it particularly effective in domains like question-answering and summarization [21,22]. This can allow for more granular metrics, if several facts are extracted and some of them are correct, allowing for a non-binary faithfulness output. This technique can detect hallucinations and factual inconsistencies by comparing extracted facts, as hallucinations and factual inconsistencies will not match with the source facts.

Fact-tuples are a structured approach for evaluating the faithfulness of generated text by extracting key facts from both the source and the output. These facts are typically represented as tuples (subject, predicate, object), capturing the essential relations and entities in the text. The methodology involves converting both the source and the generated output into sets of fact-tuples using information extraction techniques, then comparing them to measure the alignment between the source facts and the output. This comparison is commonly done using an F1 score, whereby the full list of fact-tuples available within both source and output determine the precision and recall [11].

However, this approach has the potential for incomplete or inaccurate tuple extraction, where complex or nuanced information might not be captured properly. Though, this can be mitigated somewhat through defined tuple output schema [11]. Fact-tuples also struggle with abstract or implicit knowledge, limiting their usefulness for more creative or generative tasks where direct fact alignment is less clear.

Similarly to fact-tuple extraction is Named Entity Recognition (NER), which extracts key entities from source and generated text. Discrepant entities can be assumed to either be hallucinatory or missing, and thus not fully faithful. This evaluation technique has been used to calculate precision and recall of entities to assess faithfulness and demonstrated performance superior to that of n-gram approaches [23,[24].

*Graph-based*

The use of graphs to assess faithfulness can be achieved through the translation of textual data into graph-based data. This is commonly performed using Abstract Meaning Representation (AMR). AMR is a technique that converts sentences into structured graphs, where nodes represent concepts and edges represent relationships between those concepts. For faithfulness evaluation, this graph-based methodology involves constructing AMR graphs for both the source and generated text, then comparing these graphs to assess the alignment between the underlying meaning of the two texts. This graph-based approach captures deeper semantic structures, allowing for an evaluation of whether the key concepts and relationships in the source are preserved in the generated output [25].

This approach allows for the assessment of both individual facts as well as the broader semantic and relational structure of the text, providing a more comprehensive view of faithfulness. AMR-based evaluation can detect omissions, additions, or misrepresentations of important concepts.

However, this technique can be complex and incur significant computational costs when constructing and comparing graphs. Additionally, this methodology fails to evaluate faithfulness reliably when AMR is incorrectly parsed, which can occur for ambiguous or complex language.

Additionally, small differences in graph structure may not always correspond to meaningful differences in faithfulness, leading to potential noise in the evaluation.

*LLM Evaluators*

For the purposes of this review, a metric is LLM-based if it utilizes a transformer-based architecture, with the exception of embedding-based techniques whereby whilst they typically use a pre-trained LLM to assess similarity, it is distinct from a generation task.

The methodologies differ based upon the domain in use, when language models are utilized as evaluators for faithfulness. The way in which LLMs typically operate as an evaluator is through an entailment approach. Entailment-based approaches to evaluate faithfulness utilize language models to determine whether generated text entails the source or reference text. These classifications are typically grouped into entailment, contradiction or neutral, though other studies have desired other outputs such as with a Likert 1-5 scale [26,27]. If the generated text is deemed to entail the source, it is considered factually consistent and faithful. This method can capture high-level semantic relationships between the generated and reference texts, and it has been used in various forms as a faithfulness metric across multiple domains. LLMs have been used extensively as evaluators, commonly referred to as LLM-as-a-judge, and thoroughly trialed across various domains [28].

*C) Human Evaluation*

Use of either Pearson or Spearman correlation to validate evaluation against human judgement is a common measure to validate new approaches, directly showing how well it aligns with human judgement which can be considered the gold standard for evaluating faithfulness. [11,16,26,29]

*D) Overview*

Each of the discussed metrics have various pros and cons, often defined by the data availability and the domain. These metrics are frequently used to evaluate the faithfulness of open-ended generated text. However, many of the most ubiquitous metrics are often favored for their scalability and speed but can fall short in capturing the nuances of faithfulness, especially when outputs involve abstract or complex reasoning. Even within singular domains, there is no clear, optimal metric, with various authors frequently finding different metrics to have the highest correlation with human judgement.

Many datasets used are domain-specific and less relevant in other fields, as well as many metrics being utilized for specific domains. This arises a question as to the effectiveness of domain-agnostic metrics. Furthermore, due to the unique requirements each domain has, domain-agnostic datasets are unlikely to be feasible.

IV. DATASETS FOR FAITHFULNESS EVALUATION

*A) Dataset overview*

This section will discuss the dataset requirements for each of the domains with which faithfulness is being assessed within this review, as well as common features present within these datasets.

Summarization contains the source document and summaries, where the summaries are evaluated for their adherence to the source document. These summaries can also contain labels denoting whether they are deemed faithful to the source document, typically using either a Likert scale, or a binary faithfulness measure. Furthermore, summaries can denote individual claims made from the source document, aligning this task with reading comprehension in some cases. This can result in some datasets having several unique, summaries per source, each of which requiring faithfulness evaluation. Summaries are more commonly evaluated against the source document as opposed to ground truth summaries; due to the level of ambiguity a ground truth summary can have.

Question answering datasets are required to contain QA pairs and can vary in required length of answer. There are far more QA datasets that focus on short-form or multiple-choice based questions, requiring only a single entity as the answer, or a selection from a predefined list. It is preferential for there to be multiple ground truth answers for a QA dataset, so that correct aliases can still be deemed as faithful. Multiple choice faithfulness is not within the scope of this review, which is focused on open-domain faithfulness, whereby natural text is generated to perform a task.

Machine translation datasets that have been used to evaluate the faithfulness of translations can be Quality Estimation (QE) based or can use reference translations to assess. The former only requires the original text and the generated text in order to compute a score to represent the level of similarity, whilst the latter requires a high-quality translation through which similarity between this reference and the generated translation is calculated. Additionally, human annotations of faithfulness between parallel language text have been attained within this domain, though is less common than the others mentioned – likely due to the additional requirements for the annotators to be bilingual. However, several studies have still utilized correlation with human judgement to validate faithfulness metrics within this domain.

### B) Summarization datasets

As discussed, summarization datasets all contain a source document as well as a summary (either in the form of an overview of the source, or in individual claims or questions relating to the source). There are many datasets that have been trialed in this domain, each with unique merits and desirable properties. For example, the level of abstraction is a key element in summarization datasets, with the level of concision between the source and the summary being one of the main distinctions between datasets.

In recent years, many summaries have been LLM generated and received human evaluations as to their faithfulness to the source. These human evaluations can be extremely time-consuming to produce within this domain, due to the typically long length of source documents, yet they provide the best way to benchmark other metrics. Table I details commonplace datasets within this domain for the evaluation of faithfulness.

We have made the decision to separate QA from summarization using the provision of source material – which considers reading comprehension tasks to be summarization. XSum [30] and CNN/Daily Mail [31] are two of the most commonplace datasets that have been used for evaluating summarizations, with both having been adapted by various authors to append human judgements to their summarizations

[27,29,32]. However, concerns have been raised regarding the validity of summarizations that are present within these datasets and others, leading to several authors conducting human evaluations on the summaries with respect to the source documentation, so that faithfulness can be compared against human judgement and not through the assumption that the ground truth summary is perfect [21,27,32]. This process of human annotation for faithfulness has been performed for datasets within other summarization domains as well, such as claim verification and book summarization. Both FEVER [33] and Factuality Prompts [34] provide a source document from which a claim is made, where the claim can be considered as a summary. These claims can then be labelled based on the entailment they have with the source documentation. Likewise, for book summarization there are several datasets which have relied on human annotation for judging the faithfulness of a summary [21,35,36]. Summarization within the book domain differs from the other domains discussed, largely due to the quantity of text which requires summarization – making evaluation far more problematic for both humans and computers alike.

*Dataset limitations*

Within the summarization domain, some limitations regarding the datasets have been identified resultant from this review. There are few datasets that have high quality, human produced summaries, with a large portion of datasets containing summaries that are either not representative of the source (such as XSum [30], which uses the introductory sentence as the summary) or are LLM-generated summaries which can include inaccuracies [9,27,30]. Additionally, for the assessment of faithfulness in this domain, the provision of human-labelled faithfulness judgements is not commonplace, limiting the confidence that can be placed within specific metrics and datasets.

### C) Question-Answering datasets

Datasets within the question answering domain are characterized by the presence of QA pairs, whereby a question is posed, with ground truth answer(s) provided. It is commonplace for several answers to be provided per question, due to the ubiquity with which questions have multiple correct answers, or entities have numerous valid aliases [7,37,38]. Datasets containing several answers per question can limit the rigidity with which both the exact match and lexical match methodologies operate, as well as for all additional metrics where the aliases may be closer to the outputted answer. Within the QA domain, the length of the required answer differs massively among the datasets, with the majority of QA datasets focusing on entity-based, short answers. This is likely due to the lessened ambiguity that is offered by entity-based QA, as well as the fact that commonplace metrics attain performance more comparable to humans when the answers are shorter, as metrics such as LM can attain stronger faithfulness results [7,39]. The datasets present in Table II have been used to assess faithfulness, with no external context being required for the question to be responded to (i.e. reading comprehension, or dialogues).

The majority of QA datasets that have been reviewed for this study are within the general knowledge domain, offering a wide

TABLE I

SUMMARIZATION DATASETS FOR FAITHFULNESS EVALUATION

| Author/year | Dataset | Source | Summary |
|---|---|---|---|
| Y. Kim et al., 2024 [21] | FABLES | Full books (fiction) | Model generated summaries |
| Subbiah et al., 2024 [51] | StorySumm | Reddit short stories | Model generated summaries |
| Min et al., 2023 [22] | FactScore | Model generated biographies | Sentence-level LLM extracted facts |
| Lee et al., 2022 [34] | Factuality Prompts | Statements with assigned factuality | Statements labelled for faithfulness |
| Kry´sci´ et al., 2022 [36] | BookSum | Broad genres of full books (paragraph, chapter, full) | Human and LLM summaries |
| A. Wang et al., 2022 [35] | SQuALITY | Short stories | Human and LLM summaries |
| Laban Tobias et al., 2021 [76] | SummaC | 6 summarization datasets | |
| Pagnoni et al., 2021 [32] | FRANK | CNN/DM and XSum | Model generated summaries |
| Fabbri et al., 2020 [27] | SummEval | CNN/DM | LLM CNN/DM summaries |
| Maynez et al., 2020 [29] | XSumFaith | XSum | |
| A. Wang et al., 2020 [8] | QAGS | CNN/DM and XSum | |
| Falke et al., 2019[9] | Correctness of Generated Summaries | CNN/DM | |
| Goodrich et al., 2019 [11] | WikiFact | N/A | Wikipedia sentences and fact tuples |
| Kry´sci´ski et al., 2019 [10] | FactCC | CNN/DM | CNN/DM |
| Thorne et al., 2018 [33] | FEVER | 185k Wikipedia statements with assigned factuality | |

variety of subject area. There are few datasets which contain long-form ground truth answers, with both TruthfulQA and WikiQA providing both correct and incorrect answers to a given question [39,40]. Whereas the NQ dataset provides both short-form and long-form answers that have been provided by human annotators, all of which are deemed correct [38]. Work has been conducted to provide human judgements for QA datasets, notably EVOUNA, which appends LLM answers and judgements to both TriviaQA and NQ datasets [7]. Comparably to the summarization domain, human annotation is vital to the assessment of faithfulness, ensuring that it metrics align with human judgement. However, for the QA domain, these judgements are less granular than what is seen within the summarization field, with judgements typically relating to a binary faithfulness score [7,35].

*Dataset limitations*

The main limitation within QA datasets is the range of possible answers that need to be covered, so that potential alternative correct answers, such as aliases, are included. When correct answers are not present within the dataset, any instance of the LLM providing this answer in the output will be deemed as incorrect. This limitation extends to datasets that have become outdated, with ground truth answers that are no longer temporally correct. Wang states that both NQ and TriviaQA have questions with outdated answers that should be filtered out, as well as the fact that some gold answers have inaccuracies or factual errors [7].

*D) Machine translation datasets*

Whilst there are a lot of multilingual datasets that have been gathered, primarily for training multilingual LLMs, the focus of this review is on parallel language datasets, as paired textual data can be used to assess the faithfulness of translation [41,42,43]. Table IV provides an analysis of some machine translation datasets that have been used for the assessment of faithfulness.

There are many parallel datasets for machine translation, however, comparatively few provide human annotation assessing the similarity between texts – likely due to the resources required for this level of annotation. However, small samples of human annotation are often used for validation, resultant from the high resource demands [44,45]. Due to these high resource costs to assess the similarity between multilingual text pairs, specific datasets are commonly used, notably WMT20 [46]. As despite other parallel datasets being validated with small samples of human judgement, it cannot be extrapolated to the whole dataset for accurate correlations to be attained. This issue is further exacerbated with task-oriented dialogue datasets, such as MASSIVE, requiring accurate translations for all intents and slots present [47].

*Dataset limitations*

Having native speakers or translators to post-edit automatically translated dialogues is time-consuming. On the other hand, solely based on automatically translations of

TABLE II

QUESTION ANSWERING DATASETS FOR FAITHFULNESS EVALUATION

| Author/year | Dataset | QA domain | Answer format |
|---|---|---|---|
| Mallen et al., 2023 [37] | PopQA | General knowledge | Entity-based answers |
| C. Wang et al., 2023 [7] | EVOUNA | General knowledge | Entity-based answers, LLM answers and judgements |
| S. Lin et al., 2021 [39] | TruthfulQA | Common misconceptions | Open-domain AND multiple choice |
| Min et al., 2020 [80] | AmbigQA | General knowledge | Entity-based answers |
| Kwiatkowski et al., 2019 [38] | Natural Questions | General knowledge | Entity-based and open-domain answers |
| Joshi et al., 2017 [81] | TriviaQA | General knowledge | Entity-based answers |
| Yang et al., 2015 [40] | WikiQA | General knowledge | Open-domain sentences |

English benchmark dataset produces in many cases not naturally expressed dialogues regarding the particularities of specific languages. These issues result in a lack of high quality and quantity multilingual data for evaluation and training. This issue is additionally prevalent due to web crawling being a common practice for these datasets, raising quality concerns due to the volume of data that is collected, as well as the uncertainty with which this crawled data aligns across languages.

### F) Overview

QA and summarization tasks could both likely utilize comparable datasets and schema, due to their relatively simple requirements. This can make the implementation of metrics that have demonstrated strong performance within either domain, potentially useful within the other. However, machine translation has wildly different requirements, which make this task largely incompatible with the other domains, limiting the potential application of metrics across other domains within this task. Despite these differences, there are human judgements relating to faithfulness in all the assessed tasks, allowing for metrics to be evaluated using a comparable baseline.

## V. FAITHFULNESS EVALUATION IN DIFFERENT DOMAINS

This section focuses on the evaluation methodologies that have been utilized across the aforementioned domains, discussing the best approaches that have been found. It is worth noting that there is some overlap between domains, specifically QA and summarization, with [35] considering summarization akin to long-form QA. Several studies have performed comparable work whereby summaries are segmented into individual claims, which can be evaluated in a QA domain [21,22]. Additionally, claim-based studies, whereby claims are made (these can be considered as questions) regarding source documents are both summarization and QA. Whereby, providing context to a QA evaluator, is essentially the provision of a source document, and the claim is the question [34,48].

### A) Summarization

The summarization domain has been extensively researched in with regards to faithfulness evaluation. Due to the typically long-length of summaries, and the ambiguity that this domain necessitates, LLMs have become one of the most commonplace evaluation metrics.

For the purposes of compartmentalizing the scope of this review, a task is deemed to be summarization when the context is provided to the model for the purposes of answering the question (which is typically for a summary of the context). Thus, there is some overlap between reading comprehension tasks and summarization tasks. Table IV outlines studies within the summarization domain that have performed correlation against human judgement, stating the best automated metric that they find. This table deems metrics as LLMs if they utilize text generation provided by a language model, such as BEM (BERT matching) as it uses an LLM trained to predict semantic equivalence using question, ground truth and output. All studies displayed in Table IV determine the best automated metric through correlation with human judgement, though the specific measure of correlation may differ. The findings from these authors are diverse, with LLMs, QA, embeddings and graph approaches all displaying the highest correlations for different studies. However, several of these studies have been produced with older language models, and so some comparisons may not be fair.

The use of pretrained LLMs to assess faithfulness is a ubiquitous technique within this domain, with most studies finding this methodology to provide the strongest faithfulness performance out of tested techniques [11,32,49,50]. Zhong pretrained an LLM to assess 'relevance' between a summary and the source, finding stronger correlations with human judgement than was attained by a variety of n-gram and embedding-based approaches [50]. Similarly, Maynez trained BERT-based models to calculate entailment between document and summary – finding this approach to surpass n-grams, embedding-based and QA [29]. However, other studies have failed to find strong correlation with human judgement in the summarization domain, with Fu evaluating a variety of LLM evaluators on CNN/DM and XSum (FRANK), finding minimal correlations with human judgement and stating that there are fundamental limitations in the ability of current LLMs to assess faithfulness and factuality [26].

Other studies have deviated from the generation of summaries in paragraph form and evaluated the faithfulness of LLMs through fact extraction techniques, typically where

### TABLE III
### MACHINE TRANSLATION DATASETS FOR FAITHFULNESS EVALUATION

| Author/year | Dataset | Languages | Validation | Target variable |
|---|---|---|---|---|
| FitzGerald et al., 2022 [47] | MASSIVE | 51 languages | Human translation and validation | 3-point faithfulness annotation, for matching intent and slots |
| Ramesh et al., 2022 [44] | Samanantar | 12 languages | Human evaluated subset | 5-point faithfulness |
| Kocmi et al., 2021 [74] | ToShipOrNotToShip | 101 languages | Human translation | 100-point faithfulness |
| Barrault et al., 2020 [46] | WMT20 | 19 languages (different domains and pairings) | Parallel corpus crawls – automated cleaning and human annotation | 100-point faithfulness |
| Boito et al., 2019 [45] | MaSS | 8 languages | Speech to text using translated audiobook | 5-point faithfulness |
| Cer et al., 2017 [77] | STS-B | English | SNLI corpus [78] | 5-point faithfulness |
| Steinberger et al., 2012[79] | DGT-TM | 24 EU languages | Human translation | N/A |

| Author year | Dataset | Evaluated metrics | Best automated metric |
|---|---|---|---|
| Kamalloo et al., 2023 [49] | SQuAD | N-gram, embedding, LLM | LLM |
| Ribeiro et al., 2022 [52] | CNN/DM | N-gram, QA, LLM, QA-LLM, graph, embedding | Graph |
| Ribeiro et al., 2022 [52] | XSum | N-gram, QA, LLM, graph, embedding | Graph |
| Zhong et al., 2022 [50] | SummEval | N-gram, embedding, LLM | LLM Eval |
| Pagnoni et al., 2021 [32] | CNN/Daily Mail (FRANK) | N-gram, fact-tuple embedding EM, QA, LLM, human | LLM Eval |
| Pagnoni et al., 2021 [32] | XSum (FRANK) | N-gram, QA, LLM, fact-tuple, embedding, EM | Embedding |
| Durmus et al., 2020 [16] | CNN/Daily Mail | N-gram, embedding, QA, LLM | QA |
| Durmus et al., 2020 [16] | XSum | N-gram, embedding, QA, LLM | QA |
| Maynez et al., 2020 [29] | XSum | N-gram, embedding, QA, LLM | LLM |
| A. Wang et al., 2020 [8] | CNN/DM (LLM summaries) | N-gram, embedding, QA | QA |
| A. Wang et al., 2020 [8] | XSum (LLM summaries) | N-gram, embedding, QA | QA |
| Goodrich et al., 2019 [11] | WikiFact | N-gram, fact-based, LLM fact-based | LLM fact-based |

extracted facts are claims made within the summary document and are in the form of sentences. Min only evaluates LLM-based metrics but shows that they can highly align with human judgement when atomic facts are extracted from summaries and evaluated [22]. Kim and Subbiah also use this approach to make summarization more granular and use the individual claims to evaluate faithfulness, but within book summaries and find greater correlation with human judgment when each claim is assessed [21,51]. Additionally, Tang use fact sentence extraction techniques across a range of summarization datasets and find strong performance, when using binary accuracy as opposed to correlation. Similarly, the approach taken by Ribeiro uses a graph-based faithfulness methodology akin to factual extraction through the use of AMR [52]. It is worth noting that fact extraction techniques have improved drastically since many of the discussed studies assessed their utility for faithfulness evaluation, and so greater correlations may now be able to be achieved [53,54].

Despite several studies finding LLM evaluators to have the strongest correlation with human judgement, some authors still state that this is not sufficient for specific tasks. For example, Kim finds that when evaluating summarization for a whole book that LLMs are not as capable [21]. Furthermore, they find that LLM evaluators are prone to false negatives, with far greater performance achieved when identifying faithful statements compared to unfaithful ones and this is further corroborated within Lee's study [21,34].

In addition to these binary faithfulness measures, various studies have assessed faithfulness with greater levels of granularity. Fu and Tsvilodub used a range of LLMs to assess factuality, requiring the output to be on a scale between 1 and 5, whereby 1 is completely unfaithful to the source document, and 5 is complete faithfulness [26,55]. Although, both studies found minimal statistical significance, correlation or consistency between results. However, the work conducted by Zhong evaluated the 'relevance' of a summarization on a continuous scale from 0-1 and find strong correlations with human judgement [50].

*Open issues*

It has been posited that human evaluation of long-form summaries are not reproducible [56]. This is an unsurprising observation, as summaries that are derived from long-form text (such as books) will have a far greater pool of possible valid summaries, as well as inherently being more challenging to faithfully evaluate due to the length. This limitation on summarization evaluation can lead to issues with validating faithfulness metrics within this domain, further increasing the need for reliable automated evaluation techniques. Comparably, it is due to this ambiguity that is prevalent within the summarization domain that many studies prioritize the use of source documentation to evaluate summaries against as opposed to ground truth summaries [22,29,50]. Finally, the suitability of LLMs at identifying faithful summaries is inconsistent across studies, with some authors finding performance that is comparable to human judgement, whilst others finding no statistical significance at all, highlighting the challenges and ambiguity that is contained within this domain

*B) Question-Answering*

Question answering is another common domain for faithfulness evaluation and can cover questions which necessitate open-domain long-form answers, or simple entity-based answers. For simpler, entity-based QA tasks, EM is a common metric to use due to the unambiguity it offers [7]. Whilst for more long-form answers, LM is more commonly used as it affords more lenience within the answers. However, within recent years, LLM evaluators have become far more prevalent in faithfulness evaluation within QA, typically providing an LLM evaluator with the question, ground truth and the generated output, so that the judgement can be made. Table V compares some studies that have evaluated the faithfulness of question-answering datasets using human judgement as a benchmark.

| Author, year | Dataset | Faithfulness Metrics | Best automated metric | Metric measurement |
|---|---|---|---|---|
| Yao & Barbosa, 2024 [14] | EVOUNA Natural Questions | Lexical match, embedding, LLM | LLM | Accuracy/F1 |
| Yao & Barbosa, 2024 [14] | EVOUNA TriviaQA | Lexical match, embedding, LLM | LLM | Accuracy/F1 |
| Adlakha et al., 2023 [12] | NQ-open, HotpotQA, TopiOCQA | N-gram, QA, embedding, LLM | LLM | Human judgement correlation |
| C. Wang et al., 2023 [7] | EVOUNA Natural Questions | Lexical match, embedding, LLM | Lexical match and LLM | Accuracy/F1 |

Within the QA domain, EM and LM approaches are commonplace in the evaluation of faithfulness, yet issues have been raised regarding their underestimation of model performance [7,12]. Likewise, issues have been raised with LLM evaluators, with some studies noting that performance decreases as answer-length increases [7]. As well as the observations that EM approaches typically underestimate faithfulness, whilst LLM evaluators often overestimate faithfulness [7]. Furthermore, many studies show that N-gram and embedding-based approaches fall short of exact match and LLM-based approaches.

Within the reviewed studies that compare a range of faithfulness metrics, the use of an LLM is often observed to evaluate faithfulness the most reliably [7,12,14]. However, when assessing "correctness" and "relevance", which are directly aligned with faithfulness, Abeysinghe find LLM evaluators to have little correlation with human judgement, and they specifically note the overconfidence with which LLMs deem outputs to be faithful [20]. However, it has been noted that the capabilities of LLMs in faithfulness evaluation increase as more advanced models are released and thus greater correlations are likely to be attained in the future [7]. Additionally, it has been shown that fine-tuning LLM evaluators for faithfulness evaluation improves their capabilities within this domain [39].

However, some studies do not feature the use of ground truth data when tasking LLM evaluators to produce their judgement. When this data is implemented within the input Adlakha finds strong correlations between LLM evaluators and human evaluators with OpenAI's GPT-4, attaining their strongest correlations [12,57]. A similar study was performed by Wang, which also tasked LLMs with determining the faithfulness of generated answers in accordance with provided gold answers, finding strong correlation between GPT-3.5 determinations and humans [7] .

*Open issues*

One of the primary issues in evaluating faithfulness within a QA domain is the requirement for all possible valid answers to be contained within the ground truth, when this requirement is not met a generated output can be more faithful than the evaluation would indicate [12]. Within the summarization domain this issue is mitigated through the provision of a source document from which the faithfulness is evaluated. A similar approach has been trialed within the QA domain, through using Retrieval Augmented Generation (RAG) systems to retrieve knowledge from an external document and supplement the LLMs input with this data, though no direct comparison to the use of ground truths has been provided [58].

Adlakha posits that an additional limitation in the QA domain is the predominant use of a binary variable to represent faithfulness [12]. Stating that the use of partial marks can lead to superior evaluative performance. Similar approaches have been observed within summarization, whereby summaries can be segmented into numerous facts with each fact being assessed independently to provide an average faithfulness score [21], [22]. Additionally, Wang states that LLM evaluation is sensitive to prompt, and can often use inherent knowledge to ignore golden answers, marking faithful answers (within a wider context) that are not faithful to the golden answer as faithful [7]. Thus, to ensure that LLM evaluations have optimal performance, golden answers should be verified, however, when LLMs have been trained on incorrect data this overconfidence can still lead to unfaithfulness.

*C) Machine translation*

Within machine translation, the entirety of the task is aligned with faithfulness. This is because the only goal for this task is to faithfully reproduce text in another language. However, due to grammatical and semantic differences that occur between languages, this task is notably different to those previously discussed.

Since recent developments within LLMs, they have been further utilized within this domain, though it has been discussed that they are still under-utilized [59]. Evaluation metrics for machine translation still typically use n-gram approaches, such as, BLEU [5]. However, numerous studies have expressed limitations with these metrics, relating to an inability to identify semantic similarity beyond lexical similarity [60,61,62]. Thus, in recent years metrics have been developed that aim to identify deeper semantic relationships between text, using LLM-based approaches, such as COMET and SEScore [61,63].

In addition to these approaches which require reference text, there is also Quality Estimation (QE), which aims to evaluate the faithfulness of a translation without a reference[64]. Various QE metrics have been produced, aiming to more faithfully assess translations [65,66]. Bererd finds that LLM-based evaluation methodologies attain higher correlation with human judgement when attempting QE than other automated metrics (such as n-grams) [67].

Studies that assess the open generation of translated text are compared in Table VI for the identification of the best-performing faithfulness metrics. Within machine translation, weak performance of n-gram approaches has been demonstrated across all reviewed studies in their correlation with human judgement.

TABLE VI

MACHINE TRANSLATION FAITHFULNESS EVALUATION METRIC COMPARISON

| Author, year | Dataset | Languages | Faithfulness metrics | Best automated metric | Metric measurement |
|---|---|---|---|---|---|
| Bererd & Julien, 2023 [67] | WMT20 | Unstated | N-gram, LLM-based, Embedding-based | LLM-based (QE) | Human judgement correlation |
| S. Lee et al., 2023 [62] | STS-B | En->En | N-gram, LLM-based, Embedding-based | LLM-based | Human judgement correlation |
| Xu et al., 2022 [61] | WMT20/WMT21 | En->De Zh->En | N-gram, LLM-based, Embedding-based | LLM-based | Human judgement correlation |
| Kocmi et al., 2021 [74] | ToShipOrNotToShip | 101 languages | N-gram, LLM-based, Embedding-based | COMET (LLM) | Human judgement accuracy |

BLEURT has been found to correlate strongly with human judgement, it is based on the BERT model and pretrained to estimate the similarity between text using regression models. [62]. However, this was a monolingual task, and the model used has been trained on extensive quantities of English data, this may not be seen with other languages.

Across the reviewed studies, LLM (and LLM derived metrics such as COMET) evaluation is demonstrated to outperform many other ubiquitous techniques within this domain. Both COMET and SEScore metrics perform strongly and differ from previously assessed LLM evaluation methodologies, relying instead on LLM-based rankings and text generation probabilities respectively to determine faithfulness [61].

*Open issues*

BLEURT, as well as other metrics that require a pretrained language model, have been shown to be capable evaluators in this domain. However, they are reliant on large amounts of training data and thus low-resource languages which do not have sufficient training data can lead to misguided results. Additionally, the reliable data collection of parallel linguistic corpora is a roadblock for evaluation within this domain.

## VI. CHALLENGES AND OPEN PROBLEMS

### A) Mitigating hallucination in LLMs

Through the evaluation of faithfulness in LLM text generation, many studies have used their evaluation protocols to mitigate the presence of hallucinations, and thus, increase the faithfulness of LLMs in their output. Some of these approaches are ubiquitous across domains, such as the use of external knowledge sources to ground the LLM in fact [58], [68], whilst others have shown promise in specific domains and not been trialed within others.

External knowledge sources have been utilized heavily for the mitigation of hallucinations across several domains, such as question-answering and knowledge-grounded dialogue evaluation [58,68,69]. Alinejad uses a RAG-based approach, whereby LLMs are used to answer questions when provided with either gold context or retrieved context. These responses are then fed into an LLM evaluator to determine if they are aligned [58]. Thus, this approach can replace the need for gold answers in QA-based faithfulness evaluation, mitigating the major issue of ground truth quality, whilst also aiding in reliability. However, evaluation against human judgement would be necessary for further validation of this approach.

Several studies have utilized prompting pipelines, allowing for the LLM to self-correct an unfaithful generation [70,71]. This technique has been shown to improve the faithfulness of text generation in LLMs, and when operating in conjunction with accurate faithfulness evaluation could be used to mitigate hallucinations In addition to this prompting feedback approach, it has been well documented that the prompt supplied to an LLM has major influence on the quality of the generation, irrespective of whether a refinement generation loop is used [72] . Additionally, it has been shown that outdated transformer architectures can achieve strong entailment correlations with human judgment [34]. Yet, when more modern transformer architectures are used, significant entailment improvements can be seen, with Subbiah finding a 54% increase in faithfulness evaluation between subsequent LLM architectures [51]. Thus, as the technology continues to develop, faithfulness evaluation should become ever more accurate, leading to better application of techniques such as re-prompting

However, some studies have compared the effectiveness of fine-tuning against few-shot prompting, whereby examples of desired LLM outputs are provided, and have found comparable performance with both strategies [49]. Due to the significantly reduced requirements (both computational and time-based) a few-shot prompting strategy is likely superior for most use cases, yet both should aid in mitigating hallucinations.

LLM architecture is a major factor in the prominence of hallucinations within their outputs. It has largely been shown that more modern, and larger, LLMs hallucinate less [13], [73]. Thus, hallucinations are likely to be intrinsically mitigated over time as LLMs further develop.

### B) Limits of current metrics

Currently, one of the most widely used faithfulness evaluation measures across domains is LLM evaluation [7], [12,50]. This technique has shown comparable performance in its evaluation efficacy to human judgement, and due to its versatile nature can be easily improved and applied in diverse domains. It has been demonstrated that LLM evaluation capabilities are correlated with the broad model capabilities, with more recent, larger models being shown to be more closely correlated with human judgement [13,]. However, certain drawbacks to LLM evaluation have been acknowledged across various studies. Most notably is their tendency to assume faithfulness when tasked with evaluation, evidenced through the assignment of higher scores for faithful text than unfaithful text [7]. This contrasts EM, which tends towards the classification of unfaithfulness [7]. Furthermore, when LLM evaluators provide a faithfulness score, it has been shown that the prompt used has major impact on the range of scores offered, in addition to the model used [72]. This lack of repeatability with LLM evaluators is problematic in the fair assessment of faithfulness.

### C) Future directions

There are several techniques that have been applied within the summarization domain to improve faithfulness evaluation that have not been trialed within long-form QA yet could provide similar benefits. Data structuring and fact extraction techniques have been used to simplify summarized content and reduce the ambiguity [11,21,22]. To the best of our knowledge, these approaches have not been attempted in a QA domain. Through defining a fact as a subject-relationship-object tuple, QA datasets can have ambiguity lessened, which would directly improve the exact match and lexical match approaches. This was shown by Goodrich within textual summarization and has been incorporated into QA datasets as performed by Mallen, yet no thorough analysis on this technique has been applied for QA [11,37].

Some studies have shown LLM-based metrics to better correlate with human judgement than more traditional metrics within the assessment of multilingual translation [61,62,74]. However, LLM evaluators within this field typically have skewed training data, due to the prevalence of English textual data, which can result in inconsistent evaluation capabilities across languages [62,74]. Thus, a thorough evaluation assessing the limitations of these metrics for low resource languages would aid in the validation of these approaches.

Due to the unique limitations that are apparent across all assessed metrics, the development of a fused metric that aims to mitigate the weaknesses prevalent across independent metrics has the potential to correlate more strongly with human judgment across the various domains. This can be evidenced through the exact match and LLM evaluation metrics having contrasting limitations, high rates of false negatives and false positives respectively [7].However, the fused metrics would likely need to be domain-specific due to the intricacies that each domain feature.

Data structuring is a technique that has been demonstrated to improve the evaluation of faithfulness, primarily in the summarization domain, with the segmentation of text into extracted facts bolstering performance [11,21,22]. Similar techniques have been shown within dialogues, with several studies improving entailment accuracy and consistency upon the use of some level of fact extraction [73,75]. Claim extraction-based faithfulness evaluation has been shown to be highly correlated with human judgement, surpassing other techniques used to improve these capabilities such as chain of thought prompting [51]. Data structuring and extraction is broad, and can be represented differently across domains, with some studies extracting fact tuples and others extracting sentences, yet both approaches convey utility. However, to the best of our knowledge, these techniques have not been applied in other domains yet could be utilized and potentially improve the evaluation of faithfulness. Furthermore, many studies relating to the extraction of fact-tuples could potentially benefit from the utilization of more modern extraction models, such as UniRel or REBEL, which could further improve consistency [53,54].

Due to the acknowledged limitations of EM; with a high false positive rate and LLM evaluators; with a high false negative rate, the two could be utilized in conjunction with one another as a weighted metric in the aim of mitigating the weaknesses of both.

## VIII. CONCLUSION

In this review several domains are assessed for the respective approaches that have been taken to measure faithfulness, finding that there are many metrics unique to the intricacies posed within each domain. However, we find that there is a lack of research performed in extending some of these metrics to new domains, where superior benchmarking capabilities could be achieved. This is most notable between summarization and question-answering, whereby the data structuring approaches (which typically extract claims present in the source and have improved faithfulness evaluation) used within summarization studies could be migrated to question-answering tasks. The domains that have assessed LLMs for their faithfulness evaluation capabilities have predominantly found that they are the metric that most closely aligns with human judgements. However, issues with LLM evaluators have been raised, most notably their tendency for false positives. In domains where lexical matching strategies are applicable a combinatorial metric could be utilized, mitigating the weaknesses of both LLMs and lexical matching for this task. Finally, many methodologies have been trialed to aid in mitigating the hallucinations that LLM generated text commonly suffers from, with the integration of an external knowledgebase (such as retrieval augmented generation) being both ubiquitous and successful in this task. Although other approaches exist which demonstrate an increase in the faithfulness of text generations, such as prompting frameworks which have also evidenced an increased faithfulness of outputs. It is through faithfulness evaluation that these techniques can be demonstrated to mitigate hallucinations, improving the suitability of LLMs for safety-critical scenarios, as well as to benchmark LLMs and validate their improvement for these use cases.